\documentclass[letterpaper, 10 pt, conference]{ieeeconf}  

\IEEEoverridecommandlockouts                              

\usepackage{amsmath}
\usepackage{amssymb}
\usepackage{graphicx}
\usepackage{breqn}
\usepackage{subcaption}
\usepackage{xspace}
\usepackage[dvipsnames]{xcolor}
\usepackage{wrapfig}
\usepackage{relsize}

\usepackage{algorithm}
\usepackage{algcompatible}

\usepackage{hyperref}
\usepackage[capitalise]{cleveref}

\usepackage{tabularx}
\usepackage{array}
\usepackage{booktabs}

\usepackage[hang,flushmargin]{footmisc}
\usepackage{url}

\usepackage[export]{adjustbox}

\usepackage[backend=biber,
            hyperref=true,
            url=false,
            isbn=false,
            doi=false,
            backref=false,
            style=ieee,
            citestyle=numeric-comp,
            sorting=nyt,
            block=none]{biblatex}
\title{\LARGE \bf
RT-Affordance: Affordances are Versatile \\ Intermediate Representations for Robot Manipulation
}

\author{Soroush Nasiriany$^{1,2}$, Sean Kirmani$^{1}$, Tianli Ding$^{1}$, Laura Smith$^{1}$, \\
Yuke Zhu$^{2}$, Danny Driess$^{1}$, Dorsa Sadigh$^{1}$, Ted Xiao$^{1}$
\thanks{*This work was done while the first author was a student researcher at Google DeepMind. Correspondence \texttt{soroush@cs.utexas.edu}}
\thanks{$^{1}$Google DeepMind, $^{2}$The University of Austin at Texas}%
}

\usepackage{tabularx}
\newcolumntype{L}{>{\raggedright\arraybackslash}X}

\begin{document}

\maketitle
\thispagestyle{empty}
\pagestyle{empty}

\begin{abstract}
We explore how intermediate policy representations can facilitate generalization by providing guidance on how to perform manipulation tasks.
Existing representations such as language, goal images, and trajectory sketches have been shown to be helpful, but these representations either do not provide enough context or provide over-specified context that yields less robust policies.
We propose conditioning policies on affordances, which capture the pose of the robot at key stages of the task.
Affordances offer expressive yet lightweight abstractions, are easy for users to specify, and facilitate efficient learning by transferring knowledge from large internet datasets.
Our method, RT-Affordance, is a hierarchical model that first proposes an affordance plan given the task language, and then conditions the policy on this affordance plan to perform manipulation.
Our model can flexibly bridge heterogeneous sources of supervision including large web datasets and robot trajectories.
We additionally train our model on cheap-to-collect in-domain affordance images, allowing us to learn new tasks without collecting any additional costly robot trajectories.
We show on a diverse set of novel tasks how RT-Affordance exceeds the performance of existing methods by over 50\%, and we empirically demonstrate that affordances are robust to novel settings.
Videos available at \url{https://snasiriany.me/rt-affordance}
\end{abstract}
\section{INTRODUCTION}

In recent years, we have seen the rise of large pretrained models for learning robot policies. Vision-language-action (VLA) models~\cite{rt22023arxiv,kim24openvla}, pretrained with large-scale robot data on top of vision-language models (VLMs)~\cite{geminiteam2024gemini} come with the promise of generalization to new objects, scenes, and tasks.
However, VLAs are not yet reliable enough to be deployed outside of the narrow lab settings on which they are trained.
While these shortcomings can be mitigated by expanding the scope and diversity of robot datasets, this is highly resource intensive and challenging to scale.

Alternatively, there are various ways of interfacing with the policy that can potentially facilitate generalization by providing useful guidance on how to perform manipulation tasks.
Examples of these \emph{policy representations} include language specifications~\cite{zhang2024sprint,belkhale2024rth}, goal images~\cite{black2023susie}, goal sketches~\cite{sundaresan2024rtsketch}, and trajectory sketches~\cite{gu2023rttrajectory}.
These interfaces introduce mid-level abstractions that shield the policy from reasoning in a higher dimensional input space --- leading to policies that can generalize over these intermediate representations.
While one of the most common policy representations is conditioning on language, in practice most robot datasets are labeled with underspecified descriptions of the task and language conditioning does not reveal enough guidance on how to perform the task.
Alternatively, goal image-conditioned policies provide detailed spatial context about the final goal configuration of the scene.
However, goal-images are high-dimensional, which presents learning challenges due to over-specification issues~\cite{sundaresan2024rtsketch,shah2023mutex}.
Furthermore, providing goal images at evaluation time is cumbersome for human users.
This has lead to exploration of other intermediate representations --- trajectory or goal sketches~\cite{gu2023rttrajectory,sundaresan2024rtsketch}, or keypoints~\cite{yuan2024robopoint,fangandliu2024moka} --- that attempt to provide spatial plans for the policy.
While these spatial plans are informative, they still lack sufficient information for the policy on \emph{how} to manipulate --- e.g. what pose of the gripper should take when picking up a clothes hanger.

\begin{figure}
    \centering
    \includegraphics[width=1.0\linewidth]{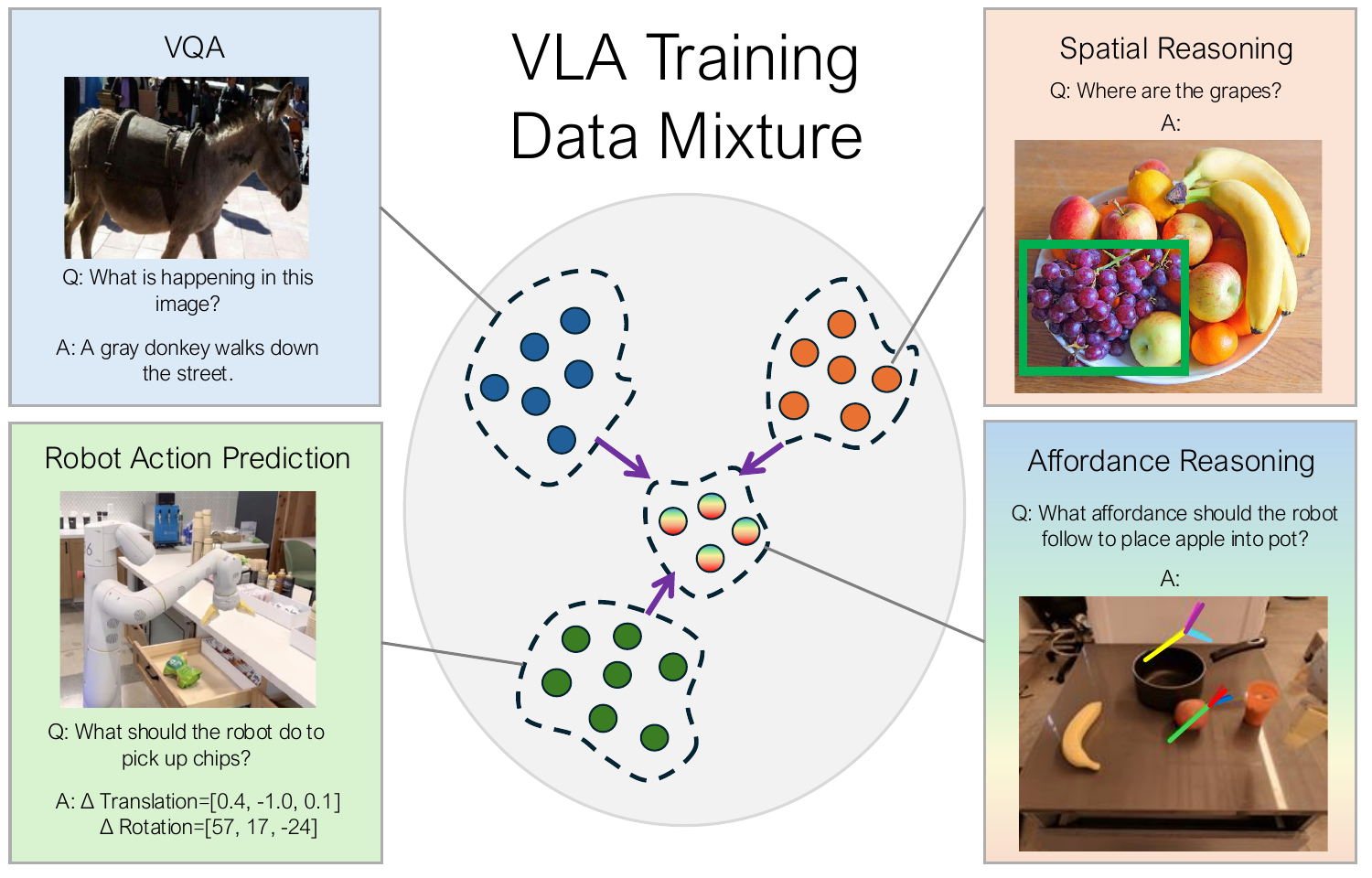}
    \caption{\footnotesize{
    \textbf{Bridging robot and internet data via affordances.} Prior work has shown the utility of co-training on robot and web datasets. However, robot actions and web content are still disjoint in their structure. We propose using affordances as a means to bridge this gap. Reasoning about affordances requires semantic and spatial reasoning, which is readily needed in VQA and spatial reasoning tasks such as object detection.
    By incorporating affordance reasoning explicitly in robot control tasks, we can better transfer knowledge from these web datasets to robot control tasks.
    }}
    \label{fig:bridging-data}
    \vspace{-3mm}
\end{figure}

\begin{figure*}
    \centering
    \includegraphics[width=1.0\linewidth]{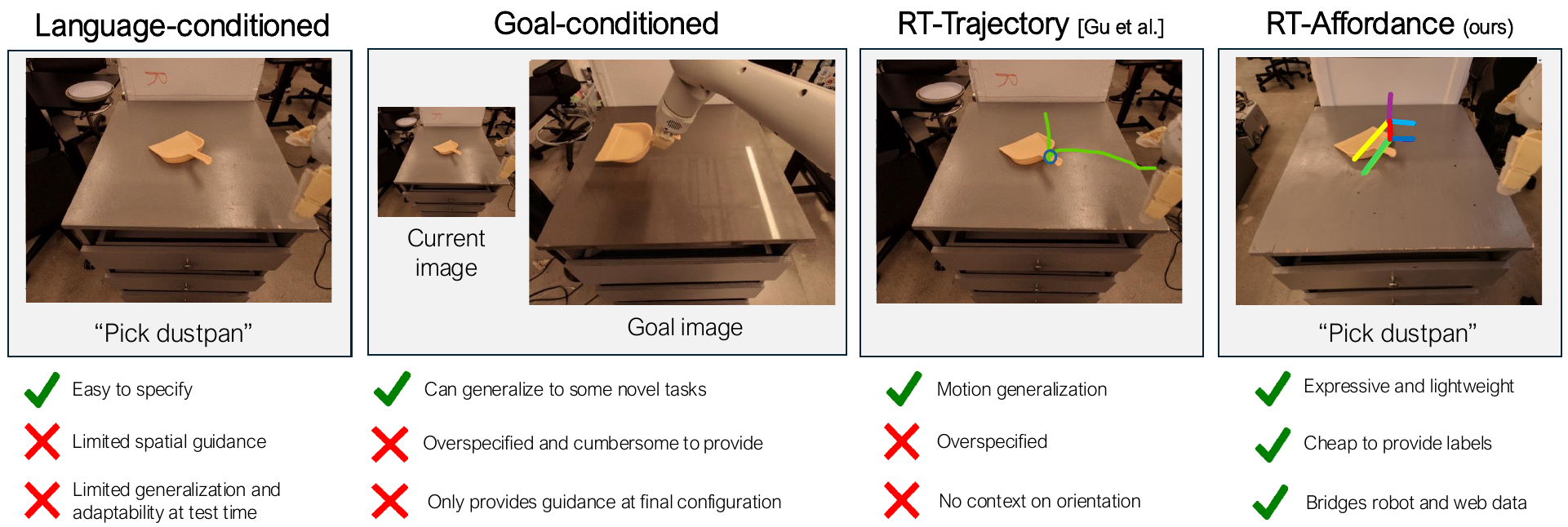}
    \caption{\footnotesize{
    \textbf{Comparison of policy interfaces.} Conditioning on language is intuitive, yet language typically does not provide enough guidance on how to perform the task. Goal images and trajectory sketches are typically over-specified and present learning challenges. We propose conditioning policies on intermediate affordance representations, which are expressive yet compact representations of tasks, making them easy to specify and to learn. 
    }}
    \label{fig:policy-interfaces}
    \vspace{-3mm}
\end{figure*}

In this work, we seek a policy representation that provides expressive yet lightweight abstractions for learning robust manipulation polices. We propose RT-Affordance, which is a policy conditioned on both language specifications and \textit{visual affordances}.
The visual affordances show the pose of the robot end effector at key stages of the task, visually projected onto the image input of the policy.
By conditioning on affordances, the robot will have access to precise yet concise guidance on how to manipulate objects.
To allow a seamless experience for the human user, we employ a hierarchical model that only requires task language from the user.
The model first predicts the affordances given a task specification in language, and then leverages the affordances as an intermediate representation to steer the policy.
The initial affordance prediction module can be trained on existing robot trajectories and web-scale datasets labeled with spatial information and affordances~\cite{Ego4D2022CVPR} (see Figure~\ref{fig:bridging-data}).
We further enhance capabilities by training on a modest dataset of cheap-to-collect in-domain images annotated with affordances. This allows us to bypass costly robot teleoperation and learn novel tasks more scalably.

We perform extensive experiments, where we show that RT-Affordance is effective across a broad range of real world tasks, achieving 69\% overall success rate compared to 15\% success rate for language-conditioned policies.
We show how incorporating both web data and cheap-to-collect affordance images allows us to learn novel tasks \textit{without collecting any additional robot demonstrations}.
Additionally, we demonstrate that the resulting affordance prediction model is robust to distribution shifts, with overall performance on out of distribution settings within 10\% of in-distribution evaluations.
\section{RELATED WORK}
\noindent \textbf{Affordances for robot manipulation.}
Affordances \cite{ardon2020affordancesrobotictasks} and grasp pose predictions have been heavily leveraged in robotics research for motion planning, grasping, and hierarchical control.
Modern data-driven methods~\cite{mousavian2019graspnet,sundermeyer2021contact} build upon prior works which leverage optimization-based approaches, and achieve performant grasp pose prediction capabilities given large-scale grasping datasets~\cite{fang2020graspnet} and point-cloud~\cite{mahler2017dex} or geometry based inductive biases~\cite{fang2023anygrasp}.
More recently, robot manipulation systems propose combining vision-language models (VLMs) with affordance or grasp prediction models and downstream control policies~\cite{huang2023voxposer,tang2023task,duan2024manipulate,huang2024rekep}.
In contrast, our method RT-Affordance does not rely on large-scale offline grasp pose specific datasets, 3D point clouds at training or test time, or simulation-based geometric planning.
\\ \\
\noindent \textbf{Learning pre-trained representations from non-action data}.
Similar to trends seen in scaling up VLMs~\cite{team2023gemini}, there has also been exploration in robotics on adapting large-scale internet data for improving perception and reasoning capabilities~\cite{driess2023palmeembodiedmultimodallanguage} which are important for downstream robot policy learning, particularly with the usage of vision-language-action (VLA) models~\cite{rt22023arxiv}.
Non-robotics interaction datasets have been particularly of interest, due to the substantial cost of real-world robotics action data such as teleoperated expert demonstrations~\cite{walke2023bridgedata,droid_2024};
representation learning methods which learn affordance prediction from internet data and human videos~\cite{damen2018epic,Ego4D2022CVPR} have been proposed~\cite{bahl2023affordances,srirama2024hrp,bharadhwaj2024track2act}.
Most similar to our method is RoboPoint~\cite{yuan2024robopoint}, which proposes fine-tuning a VLM to predict points which represent spatial affordances by leveraging procedural 3D scene generation in simulation.
Our method RT-Affordance also studies predicting spatial affordances, but proposes a more descriptive affordance representation beyond a single point, and also does not require large-scale simulated scene generation.
\\ \\
\noindent \textbf{Intermediate representations for policy conditioning}.
Prior works have studied how multi-task robot manipulation policies can be conditioned on various types of representations and interfaces to perform different manipulation skills.
Popular interfaces have included one-hot task vectors~\cite{kalashnikov2021mt}, latent skill or task embeddings~\cite{hausman2018learning,james2018task,pertsch2021accelerating}, templated or natural language~\cite{jang2022bcz,rt12022arxiv,stepputtis2020languageconditioned,zhang2024sprint,nair2021learning}, object-centric representations~\cite{jiang2023vima,moo2023arxiv,sundaresan2023kite,fangandliu2024moka}, trajectories~\cite{gu2023rttrajectory,wen2023anypoint}, goal images or sketches~\cite{bousmalis2023robocat,chebotar2021actionable,sundaresan2024rtsketch,nasiriany2019planning,black2023susie,cui2022playtopolicy,cui22can}, and videos~\cite{finn2017oneshot,dasari2020oneshot,jain2024vid2robot}.
Our method leverages affordances represented visually or textually as an interface, which strikes a balance between flexibility, expressivity, and data efficiency (see Figure~\ref{fig:policy-interfaces}).

\begin{figure*}
    \centering
    \includegraphics[width=1.0\linewidth]{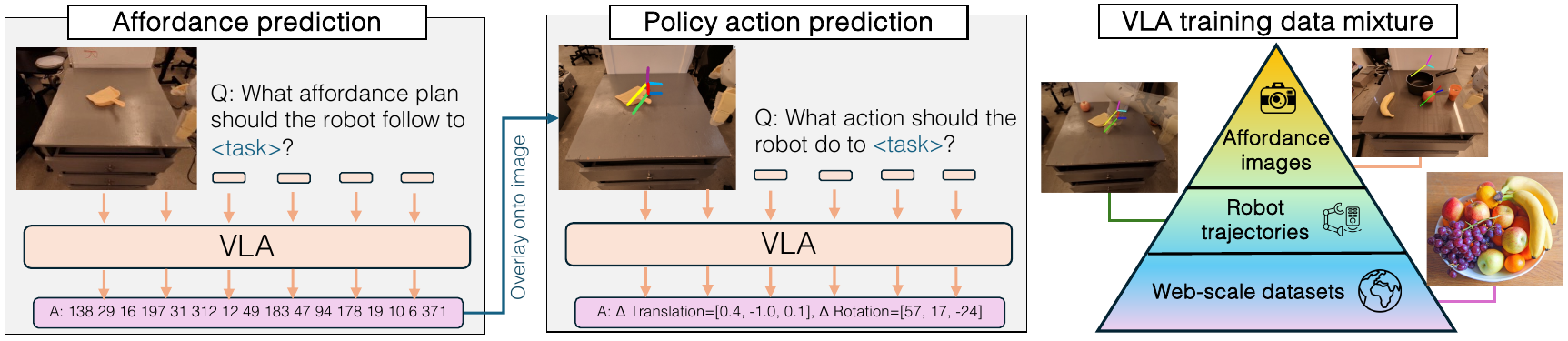}
    \caption{\footnotesize{
    \textbf{Model overview.} Our hierarchical model first predicts the affordance plan given the task language and initial image of the task.
    We overlay the affordance (pixel xy values in raw text form) onto the image, and subsequently condition the policy on images overlaid with the affordance plan.
    We co-train the model on web datasets (largest data source), robot trajectories, and a modest number of cheap-to-collect images labeled with affordances.
    }}
    \label{fig:method-overview}
    \vspace{-3mm}
\end{figure*}

\section{RT-A: Affordance-Based Policy Learning}
Our goal is to implement an intermediate policy interface that (1) is an expressive yet compact representation for a broad set of manipulation tasks,
(2) can effectively bridge knowledge from external datasets and facilitate generalization,
and (3) enables learning novel tasks through cheap, in-domain data collection.
We propose RT-Affordance (RT-A), a hierarchical policy which first proposes an affordance plan via an affordance generator, and then generates actions via an affordance-conditioned policy.
We will first introduce the affordance-conditioned policy and subsequently introduce the affordance generator.

\subsection{Affordance-conditioned policies}
\label{sec:aff-cond-policy}
We are given a dataset of robot trajectories $\mathcal{D} = \{l, \{(o_i, e_i, g_i, a_i)\}_{i=0}^T\}$;
each trajectory consists of a language instruction $l$ and a sequence of images $o_i$, actions $a_i$, end-effector poses $e_i$ and gripper states $g_i$.
We learn an affordance-conditioned policy $\pi(a | l, o, q)$ that generates actions given the language instruction $l$, current image $o$, and additionally the \textit{affordance plan} $q$.
We define the affordance plan as the sequence of robot end effector poses corresponding to key timesteps in the trajectory, $q = (e_{t_1}, e_{t_2}, ..., e_{t_n})$.
These timesteps capture critical stages in the task execution, for example when the robot is about to come in contact with objects or encounters bottleneck states.
We can employ a variety of approaches to extract these timesteps.
In practice we adopt a simple and scalable solution: we automatically extract from proprioception data timesteps when the gripper state changes from open to close ($g_{i - 1} > \alpha$ and $g_i < \alpha$ for some constant $\alpha$) or vice versa from close to open, or the final timestep of the trajectory.
This implicitly captures object-centric interactions corresponding to the stages in the task when the robot contacts, grasps, pushes, or lets go of objects.
Compared to conditioning on language as in prior work~\cite{rt22023arxiv}, the affordance plans in RT-A policies reveal precise spatial information about how to manipulate objects.
These affordance plans not only reveal the position of the robot end effector but also orientation, which is critical for fine-grained manipulation.
However, solely conditioning on affordance plans may not reveal full context about the task, and we thus opt to condition the policy on both affordance plans and language.
This ensures that we retain the full expressiveness of language-conditioned policies, while benefiting from the additional context provided by affordance plans.

We train the affordance-conditioned policy via behavioral cloning and additionally co-train on web datasets, in a similar manner as in RT-2.
We can represent these affordances either as tokenized text values passed as input to the policy, or by overlaying them onto the image using a visual operator $\psi(o, q)$, following similar techniques in prior work~\cite{nasiriany2024pivot,gu2023rttrajectory}.
In our implementation we visually project the outline of the robot hand at the poses $e_i$ onto the image.
Specifically given $e_i$ we compute the 3D positions of the leftmost end effector tip, rightmost finger tip, top of end effector, and arm, and project these points onto the 2D image.
We connect these points to make an outline. See Figure~\ref{fig:method-overview} for an illustration.
We designate unique colors to each of the affordances overlaid onto the image to capture temporal ordering.
Note that this projection step assumes knowledge of the robot camera intrinsics and extrinsics which is readily available for many robot platforms.
If this information is not available, we can opt to condition the policy on the affordance plan directly as tokenized text values.

\subsection{Learning to predict affordances}
\label{sec:aff-pred-model}

We can deploy the affordance-conditioned policy by asking the human user to provide affordance plans and language goals to the policy at inference time.
The affordance plans can be provided easily by marking them visually onto the image using a UI interface, without moving the robot of changing the scene.
Compared to prior approaches such as conditioning on goal images or trajectory sketches, affordance plans are lower dimensional, making them easier to provide.
We can also learn models to predict affordance plans automatically, sidestepping the need for humans to provide affordances at all at test time.

We learn an affordance prediction model $\phi(q | l, o)$ which predicts the affordance plan given the language task instruction $l$ and initial image of the scene $o$.
To train the model we extract $(o, l, q)$ tuples from the same robot dataset $\mathcal{D}$ used to train $\pi$ and we also co-train the model with web datasets.
In applications where we have access to camera information we predict the \textit{xy pixel locations} of the end effector points, allowing us to better transfer knowledge with other existing web datasets such as object detection.
In some applications, training on these datasets may not yield adequate performance and we may seek additional training data to further improve the capabilities of the model.
Instead of collecting additional demonstrations through expensive robot teleoperation, we can collect a set of images with corresponding task labels, ie. $\mathcal{D}_{\text{aug}} = \{(o_i, l_i)\}_{i=0}^n$.
We can collect hundreds or thousands of these images at a fraction of the cost compared to costly teleoperation.
After this data collection process we can annotate each image with the affordance plan through a posthoc lableing procedure, either manually through a UI interface or with the aid of tools such as VLMs and grasp planners, as shown in previous work~\cite{gu2023rttrajectory}.
This annotation process can be performed efficiently offline and can be crowdsourced without the need for expensive robot hardware or teleoperation.
This additional data collection process enables us to improve performance on the downstream robot task with minimal additional effort and allows us to bypass costly robot teleoperation, which we will demonstrate in our experiments.

\subsection{Model Inference}
We are given the initial image of the scene $o_{\text{init}}$ and a natural language task instruction $l$.
We can either prompt a human or the affordance prediction model $\phi(q | l, o_{\text{init}})$ to obtain the affordance plan $q$.
The affordance plan is projected onto the image, i.e., $\psi(o, q)$ and we prompt the policy $\pi(a | l, \psi(o, q))$ with the language instruction and annotated image to execute the task.
We can optionally replan updated affordance plans at fixed or adaptive intervals to handle novel scenarios that arise during the execution of the policy.
\section{EXPERIMENTS}

In our experiments we are interested in exploring the following questions:
\begin{itemize}
    \item Are affordance-conditioned polices broadly useful for performing diverse manipulation tasks?
    \item Can affordances enable efficiently learning novel tasks, without costly additional robot teleoperation?
    \item How well do affordance prediction models generalize to novel objects, camera views, and backgrounds?
\end{itemize}

\subsection{Experiment implementation}
We use the robot manipulator from RT-1~\cite{rt12022arxiv}. The arm is controlled via Cartesian end-effector control.
The robot observes the environment from a single head-mounted camera.
Our robot demonstration datasets comprise three phases of data collection: (1) the RT-1 dataset~\cite{rt12022arxiv} which focuses on basic manipulation skills, (2) the MOO dataset~\cite{moo2023arxiv} which focuses on picking diverse objects, and (3) an additional set of trajectories targeting more dexterous tasks.
We use the same web datasets from RT-2 for co-training.
We adopt PaLM-E 2~\cite{anil2023palm2technicalreport,driess2023palmeembodiedmultimodallanguage} as the underlying model and use the 1-billion parameter variant, unless otherwise noted.
We train and evaluate vision-language-action models (VLAs) which share the same underlying model but adopt different policy input interfaces.
All methods are trained on the same number of robot trajectories and same web datasets.
We train the affordance prediction model with the hindsight affordance labels from the robot trajectory datasets, in addition to a set of $\sim$750 cheap-to-collect images manually annotated with affordance labels.
We collect these images by placing diverse objects on the table in front of the robot and taking snapshots of the scene.
These images include the tasks and objects from our grasping tasks and additional tasks beyond grasping which we will outline the tasks in the following sections.
We collect all of these images in approximately one hour and dedicate an additional two hours annotating them with affordance labels afterwards.
This data collection process is significantly faster and more scalable than collecting robot trajectories.
We train a dedicated VLA for affordance prediction, trained on web data, affordances extracted from robot trajectories, and in-domain affordance images.

\subsection{Learning to grasp novel objects efficiently}
In our first experiment we investigate how affordances facilitate learning to grasp novel objects.
Grasping is a ubiquitous skill demanded across a wide range of tasks, and it is important that robots can grasp diverse objects in a robust manner.
We design a benchmark of picking diverse household objects, including dustpans, kettles, pots, boxes, and headphones.
In contrast to simple objects with rigid convex shapes, we selected these objects as they encompass complex shapes and require fine-grained part-level reasoning in order to successfully grasp them.
Note that our benchmark focuses on unseen object categories, meaning that they are not present in any of our robot trajectory datasets.
We place the object on a tabletop in addition to two or three distractor objects coming from a wide range of object categories.
We run comprehensive evaluations comparing our method to prior state-of-the-art approaches.
We evaluate each method across five rollouts per object category and record the task success rates in Table~\ref{fig:main-results-table}.

\begin{table}
\centering
\small
\setlength{\tabcolsep}{4pt}

\begin{tabular}{lcccc}
\toprule
   & RT-2 & GC-RT-2 & \textbf{RT-A} & \textbf{RT-A } \\
   & & & \textbf{(Oracle Aff)} & \textbf{(Ours)} \\
   \midrule
Pick dustpan & 1/5 & 1/5 & 3/5 & \textbf{4/5} \\
Pick kettle & 1/5 & 1/5 & \textbf{4/5} & \textbf{4/5} \\
Pick pot & 0/5 & 1/5 & \textbf{4/5} & 1/5 \\
Pick box & \textbf{4/5} & 1/5 & \textbf{4/5} & \textbf{4/5} \\
Pick headphones & 1/5 & 2/5 & \textbf{4/5} & \textbf{4/5} \\ \midrule
Average & 28\% & 24\% & \textbf{76\%} & 68\% \\ \bottomrule
\end{tabular}
\caption{\footnotesize{\textbf{Experimental results on grasping.} We compare our model to state-of-the-art VLAs, including language-conditioned and goal-conditioned policies. These methods fail to grasp objects precisely achieving success rates of under 30\%. In contrast our affordance-conditioned policy paired with oracle human-provided affordances achieves 76\% performance, and when employing an affordance prediction model to infer affordance automatically we observe a 68\% success rate.}}
\label{fig:main-results-table}
\vspace{-4mm}
\end{table}

First we compare to \textbf{RT-2}~\cite{rt22023arxiv}, a state-of-the-art language-conditioned robot policy learning model notable for its impressive capabilities in understanding novel semantic concepts and objects.
Despite these capabilities, we find that it struggles on our suite of evaluations, achieving an average success rate of just 28\%.
We observe that the policy is generally capable in identifying the correct object on the table and reaching the vicinity of the object but is unable to grasp the object at the appropriate location.
For example, the robot attempts to grasp at the center of the dustpan rather than the handle, resulting in an unsuccessful grasp.
Similarly with picking the pot the robot tries to grasp around the base of the pot rather than handle.
However, it is generally capable of picking boxes.
We also tried to prompt the policy with specific language instructions indicating how to grasp the object (e.g. ``pick the dustpan by the handle'') but the policy failed to follow these instructions effectively.

We also evaluate a goal-conditioned variant of RT-2 (\textbf{GC-RT-2}), which replaces language-conditioning for image goal-conditioning. We use the larger 24-billion variant PaLM 2 backbone to accommodate the additional goal-image passed into the policy.
We run evaluations on the same objects, and for each episode we manually take a snapshot of the robot having grasped and lifted the object in the air at the final goal configuration.
We observe an average success rate of just 24\%.
While the goal image conveys the precise pose at which to grasp the object, the policy is unable to precisely grasp the object at this pose.
One common failure mode is that the robot fails to rotate its end effector sufficiently in order to achieve the correct pregrasp orientation.

Next we compare our hindsight affordance model RT-A.
We condition the policy with the language instruction and visual affordances overlaid on top of the current image.
We first evaluate the model with oracle affordances, ie. for each trial we manually provide the pregrasp and goal affordance poses of the robot.
We call this self-baseline of our method \textbf{RT-A (Oracle Aff)}.
We observe a significant improvement in policy performance, achieving 76\% average success.
The policy is faithful in executing the human provided affordance poses, and failures are only due to small imperfections from the robot policy in following the given affordance poses.
Again, we highlight that none of these object categories are present in the robot trajectory datasets, making this a effective method for grasping a broad set of objects.

\begin{table}
\centering
\small
\setlength{\tabcolsep}{4pt}

\begin{tabular}{lccc}
\toprule
   & RT-2 & \textbf{RT-A} & \textbf{RT-A } \\
   & & \textbf{(Oracle Aff)} & \textbf{(Ours)} \\
   \midrule
Place apple into pot & 0/5 & \textbf{4/5} & 3/5 \\
Place peach onto plate & 1/5 & \textbf{4/5} & \textbf{4/5} \\
Place bell pepper into basket & 0/5 & 3/5 & \textbf{4/5} \\
Place eggplant into box & 0/5 & 2/5 & \textbf{3/5} \\
Close the cubby & 0/5 & \textbf{4/5} & \textbf{4/5} \\
Turn sink faucet & 0/5 & \textbf{4/5} & 3/5 \\
\midrule
Average & 3\% & \textbf{70\%} & \textbf{70\%} \\ \bottomrule
\end{tabular}
\caption{\footnotesize{\textbf{Beyond grasping.} RT-A is applicable to a broad set of tasks, including placing objects into various receptacles and manipulating articulated objects. RT-Affordance with a successs rate of 70\% performs significantly better than RT-2 (only 3\% success) on these tasks.}}
\label{fig:addl-tasks-table}
\vspace{-3mm}
\end{table}

Finally we compare to the full hierarchical variant of our method in which we predict affordance plans before conditioning the policy on these plans (\textbf{RT-A}).
We see an average performance of 68\%, which is close to the performance of the policy conditioned on oracle affordances.
Compared to the oracle affordance self-baseline we see similar performance across all object categories except picking the pot.
Here we observe that while the predicted affordances are reasonably placed around the handle, there are some edge cases, for example the robot freezing or picking the objects but not lifting them sufficiently in the air.
Such errors may be mitigated by perturbing or re-planning the affordances in an adaptive manner, and we leave this for future work.

\subsection{Beyond object picking}
We demonstrate that these findings are not exclusive to grasping tasks but can be extended to a range of manipulation tasks.
We compare RT-A to the next best baseline from the previous experiments, the language-conditioned RT-2 model, on an additional set of manipulation tasks.
We consider tasks involving placing objects into receptacles (eg. ``place apple into pot'', ``place bell pepper into basket''), and articulated manipulation (``close the cubby'').
Again, we highlight that these tasks are \textit{unseen} in the robot trajectory datasets and demand precise spatial reasoning and execution.
See Table~\ref{fig:addl-tasks-table} for results.
Surprisingly, the RT-2 baseline performs quite poorly in this setting achieving only 3\% success rate.
We observe a range of failures, including unreliable grasping of objects, freezing after grasping the object, placing the object next to the receptacle rather than into the receptacle.
Using the same underlying VLA architecture but additionally conditioning on affordances, and we employ the affordance conditioned model trained to predict affordances on a handful of images annotated with affordance labels.
We see a significant improvement of performance, with 70\% success rate using our affordance prediction model.
These results show that affordances are a flexible form of task specification that can describe a broad set of tasks.
In cases where the user provides oracle affordances at evaluation, we can solve novel tasks without any additional data, and training our affordance prediction model to infer affordances automatically only incurs a small budget to collect and annotate images.
In contrast, improving the capabilities of language-conditioned or goal-conditioned policies would require fine-tuning on dozens or hundreds of additional robot demonstrations collected through teleoperation~\cite{belkhale2024rth,liu2022sirius}, which is significantly more expensive and less scalable.

\begin{figure}
    \centering
    \includegraphics[width=1.0\linewidth]{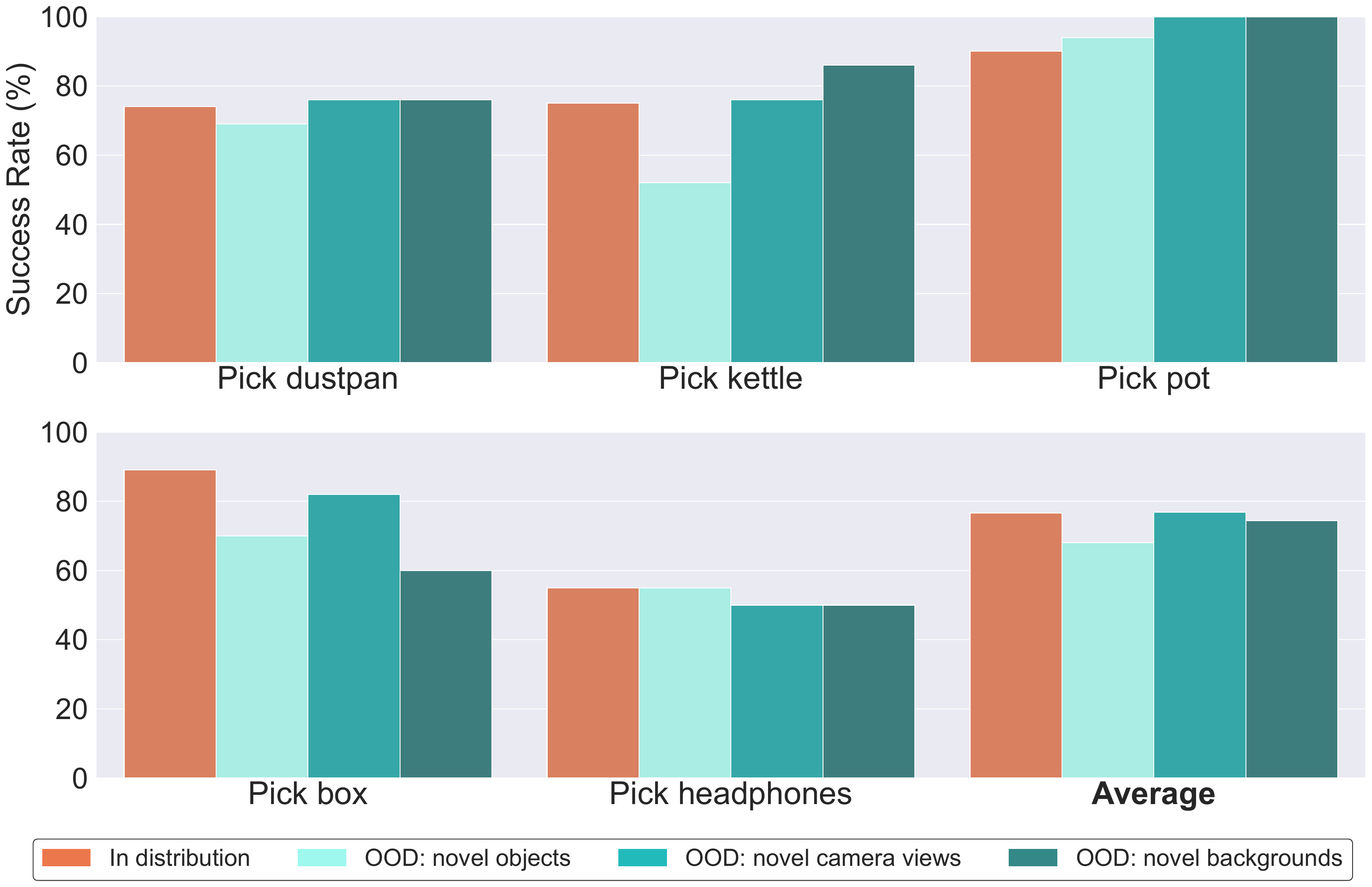}
    \caption{\footnotesize{\textbf{Evaluation of the affordance prediction model on out of distribution scenarios.} We perform a comprehensive evaluation of the affordance prediction model on in-distribution and out-of-distribution (OOD) and observe a graceful degradation of performance in OOD settings.}}
    \label{fig:offline-evals-plot}
    \vspace{-3mm}
\end{figure}

\begin{figure*}
    \centering
    \includegraphics[width=1.0\linewidth]{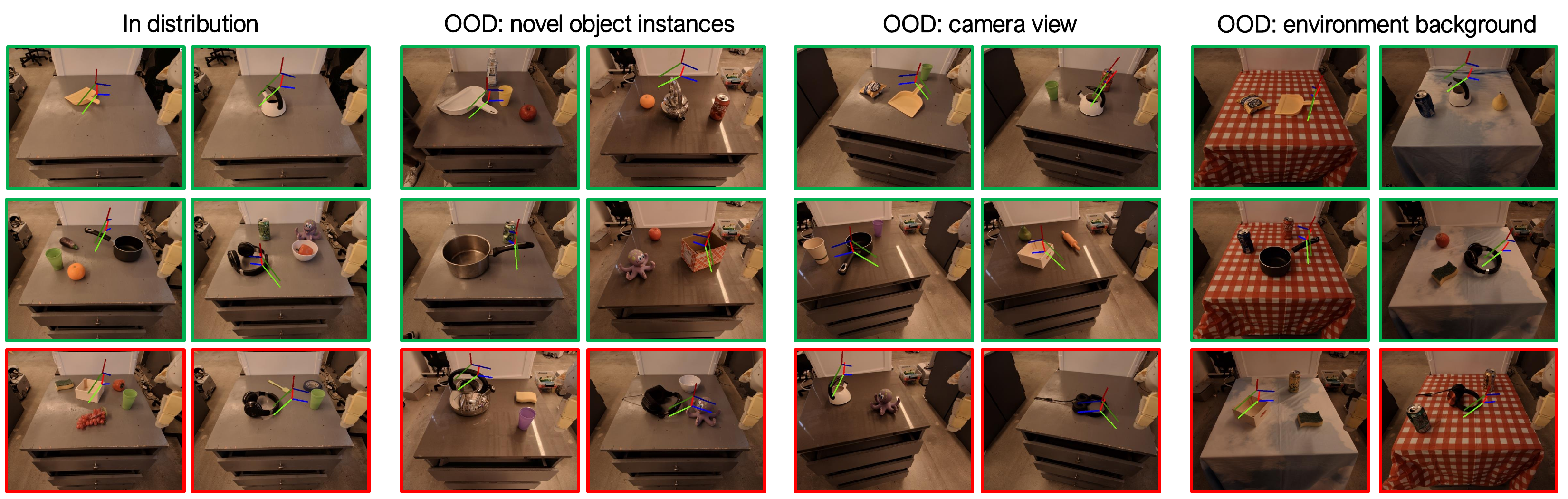}
    \caption{\footnotesize{
    \textbf{Robustness to out of distribution factors} We show examples of successful and incorrect predictions of our affordance prediction model across in-distribution and out-of-distribution settings. Successful predictions are highlighted in green and incorrect predictions are highlighted in red.
    }}
    \label{fig:ood-evals-vis}
    \vspace{-3mm}
\end{figure*}

\subsection{Robustness to out of distribution factors}
\label{sec:ood}
Next, we perform an analysis of the affordance prediction model.
In order for the affordance prediction model to be useful it needs to be robust to a wide range of out-of-distribution (OOD) settings.
To better understand this, we perform a comprehensive evaluation on the grasping tasks from Table~\ref{fig:main-results-table} comparing the following settings:
\begin{itemize}
    \item \textbf{In-distribution}: evaluating the model under the same distribution it was trained on. ie. same object instances, same camera view, and same environment background.
    \item \textbf{OOD: novel objects}: evaluating the model with novel object instances on which it was not trained on. 
    \item \textbf{OOD: novel camera view}: evaluating the model with images taken with significant camera shift.
    \item \textbf{OOD: novel background}: evaluating the model with novel object textures.
\end{itemize}

\begin{table}
\centering
\small
\setlength{\tabcolsep}{4pt}

\begin{tabular}{lccc}
\toprule
   & \textbf{Ours} & \textbf{No aug data} & \textbf{No web data} \\
   \midrule
Pick dustpan & \textbf{74\%} & 20\% & 3\% \\
Pick kettle & \textbf{75\%} & 30\% & 10\% \\
Pick pot & \textbf{90\%} & 10\% & 14\% \\
Pick box & \textbf{89\%} & 33\% & 11\% \\
Pick headphones & \textbf{55\%} & 28\% & 16\% \\ \midrule
Average & \textbf{77\%} & 24\% & 11\% \\ \bottomrule
\end{tabular}
\caption{\footnotesize{\textbf{Ablation study.} We perform an ablation study of our affordance prediction model the same in-distribution evaluations as Figure~\ref{fig:offline-evals-plot}. We find that removing the augmented dataset of affordance images significantly diminishes performance, and removing web datasets for co-training diminishes performance even further.}}
\label{table:ablation-evals-table}
\vspace{-4mm}
\end{table}

We perform a comprehensive offline human evaluation over hundreds of test images, where for each image we assess whether the model's predicted affordance would result in a successful grasp, assuming that the policy can follow the given affordances perfectly.
We report the results in Figure~\ref{fig:offline-evals-plot}.
First, we see that the affordance prediction model is general capable in in-distribution settings, with 77\% of trials classified as success.
Across the OOD settings model performance degrades gracefully, falling no more than 10\% compared to the in-distribution setting.
Some factors affect model performance more than others.
With novel camera views the performance is nearly identical at 77\%, and with novel backgrounds performance only falls at 3\% on average.
However with novel object instances the performance drops the most, especially for grasping novel instances of kettles and boxes.
We provide illustrative examples in Figure~\ref{fig:ood-evals-vis}.

\subsection{Ablation study}
We perform an ablation study on our affordance prediction model, where we study the impact of different data sources on the model. Our model is trained on the full data mixture including (1) robot trajectories, (2) web datasets, and (3) the ~750 additional augmented affordance images we collected.
We perform ablations where we (a) exclude the augmented data (\textbf{No aug data)} and (b) exclude web datasets (\textbf{No web data}).
We compare these settings on the same in-distribution evaluation suite outlined in Section~\ref{sec:ood}, and we report results in Table~\ref{table:ablation-evals-table}.
We see that removing these sources of data leads to a large drop in performance.
We hypothesize that large web datasets play an important role for training robust models, and that our augmented data is needed to train performant models for specific downstream tasks.

\section{CONCLUSION}
We have presented RT-Affordance, a hierarchical method that uses affordances as an intermediate representation for policies.
Affordances provide precise spatial guidance on how to preform a manipulation task, and they are easy for humans to specify.
We have shown empirically that affordance-conditioned policies can perform a wide range of novel tasks without requiring additional human demonstrations.
Additionally, we have shown how we can learn models to predict affordances during deployment using cheap-to-collect images, and that these models are robust.
One limitation that we observed is that our policy did not exhibit generalization to completely novel motions or skills.
This has been noted in other VLA works~\cite{rt22023arxiv} as well, and we are interested in exploring this in future work.
In addition we are interested in exploring the complementary strengths of different policy representations and combining their capabilities into a single model that can share knowledge across representations.


\section*{ACKNOWLEDGMENTS}
We thank Fei Xia for fruitful discussions and Dhruv Shah for providing feedback on writing.
We also thank Justice Carjabal, Tomas Jackson, and other robot operators at Google DeepMind for assisting us in data collection and evaluation.

\newpage
\printbibliography 
\end{document}